\documentclass[preprint,12pt]{elsarticle}

\usepackage{amssymb}
\usepackage{amsmath}

\usepackage{lineno}
\usepackage{enumitem}

\usepackage{float}

\usepackage{multirow}
\usepackage{array}

\newcolumntype{C}{>{\centering\arraybackslash}m{3cm}}
\newcolumntype{L}{>{\raggedright\arraybackslash}m{5cm}}
\newcolumntype{R}{>{\raggedleft\arraybackslash}m{3cm}}

\newcommand{\coltwofont}{\small}
\newcommand{\colthreefont}{\small}

\journal{Chaos, Solitons and Fractals}

\begin{document}

\begin{frontmatter}



\title{Self-Training the Neurochaos Learning Algorithm} 


\author[label1]{Anusree M} 
\author[label1]{Akhila Henry}
\author[label1]{Pramod P Nair\corref{cor1}}

\cortext[cor1]{Corresponding Author, Email: {pramodpn@am.amrita.edu}}

\affiliation[label1]{organization={Department of Mathematics, Amrita Vishwa Vidyapeetham},
            city={Amritapuri},
            country={India}}

\begin{abstract}

In numerous practical applications, acquiring substantial quantities of labelled data is challenging and expensive, but unlabelled data is readily accessible. Conventional supervised learning methods frequently underperform in scenarios characterised by little labelled data or imbalanced datasets. This study introduces a hybrid semi-supervised learning (SSL) architecture that integrates Neurochaos Learning (NL) with a threshold-based Self-Training (ST) method to overcome this constraint. The NL architecture converts input characteristics into chaos-based firing-rate representations that encapsulate nonlinear relationships within the data, whereas ST progressively enlarges the labelled set utilising high-confidence pseudo-labelled samples. The model's performance is assessed using ten benchmark datasets and five machine learning classifiers, with 85\% of the training data considered unlabelled and just 15\% utilised as labelled data. The proposed Self-Training Neurochaos Learning (NL+ST) architecture consistently attains superior performance gain relative to standalone ST models, especially on limited, nonlinear and imbalanced datasets like \textit{Iris}(188.66 \%), \textit{Wine}(158.58 \%) and \textit{Glass Identification}(110.48 \%). The results indicate that using chaos-based feature extraction with SSL improves generalisation, resilience, and classification accuracy in low-data contexts.
\end{abstract}


\end{frontmatter}


\section{Introduction}
\label{sec1}

In numerous real-world scenarios, we frequently possess an extensive dataset, although only a limited fraction is labelled. Data labelling is labour-intensive, costly, and occasionally demands specialised expertise. Nonetheless, the unlabelled data is readily accessible and encompasses valuable insights regarding the dataset's structure and patterns. Acquiring knowledge from both labelled and unlabelled data enhances model performance and generalisation. Utilising the plentiful unlabelled data enables the development of superior models, even in the presence of little labelled data. This methodology is extensively employed in domains such as text classification \citep{duarte2023review}, where substantial quantities of unlabelled text data are accessible. Sentiment analysis on social media \citep{silva2016survey} and web content categorisation \citep{blum1998combining} also get advantages from utilising unlabelled data.

Semi-supervised learning (SSL) methodologies are employed in scenarios where a limited quantity of labelled data exists with a substantial volume of unlabelled data \citep{van2020survey}. These strategies enhance model learning by integrating the advantages of both supervised and unsupervised learning. Various SSL techniques exist. Each methodology employs distinct methods to utilise unlabelled data for enhancing learning. Nonetheless, these algorithms exhibit several drawbacks, including susceptibility to erroneous pseudo-labels and challenges in managing noisy or imbalanced data. To address these problems, numerous research investigations are being conducted to build more robust and efficient SSL methodologies.

Neurochaos Learning (NL) is a method for learning that draws inspiration from chaos theory and neural computation principles \citep{harikrishnan2021noise}. It employs chaotic dynamics to convert input data into a more complex and distinguishable feature space. This enables the model to identify intricate, nonlinear patterns within the data. The algorithm functions proficiently even with noisy and imbalanced datasets and operates effectively with minimal labelled data \citep{anusree2025understanding}. It is straightforward, effective, and necessitates minimal instruction. Two primary categories of NL architectures exist: one employs a cosine similarity-based classifier \citep{balakrishnan2019chaosnet}, while the other integrates chaotic transformation with conventional machine learning classifiers to enhance performance \citep{sethi2023neurochaos}.

This study seeks to broaden the utilisation of NL from supervised learning tasks to SSL challenges through a Self-Training Neurochaos Learning (NL+ST) method. It employs the singular chaotic characteristic known as the firing rate, obtained from the chaotic transformation, as the primary feature for learning. This feature is subsequently employed to self-train various classic machine learning algorithms for categorisation, utilising the readily available unlabelled data. This methodology enabled us to integrate the advantages of chaotic approaches and the ST method. This combination mitigates sensitivity to noisy pseudo-labels and enhances generalisation from limited labelled data.

The rest of this paper is organized as follows. Section~\ref{sec:literature} presents a detailed literature review of existing NL architectures, SSL techniques, ST algorithms and related works. Section~\ref{sec:methodology} explains the proposed hybrid SSL methodology that combines NL with threshold-based ST. Section~\ref{sec:results} discusses the experimental results and performance comparisons obtained using different machine learning classifiers. Finally, Section~\ref{sec:conclusion} provides the conclusion of this study and outlines possible directions for future work.


\section{Literature Review}
\label{sec:literature}

SSL is a machine learning methodology that utilises both labelled and unlabelled data. SSL facilitates the utilisation of extensive unlabelled data in conjunction with a limited labelled dataset to enhance model performance. It operates under numerous fundamental assumptions, including smoothness, low-density separation, and the manifold assumption. These assumptions assert that data points in proximity or situated on the same low-dimensional manifold are likely to share identical labels. Numerous algorithms have been formulated within this domain, including self-training (ST) \citep{amini2025self}, co-training \citep{blum1998combining}, tri-training \citep{zhou2005tri}, graph-based methodologies \citep{song2022graph}, and semi-supervised adaptations of conventional models such as SVMs \citep{bennett1998semi} and boosting algorithms \citep{mallapragada2008semiboost}. These algorithms integrate information from both labelled and unlabelled data to enhance generalisation and accuracy.

SSL has been effectively implemented in numerous practical applications. It is extensively utilised in medical image analysis \citep{jin2025advancements}, where expert-annotated data is scarce, yet substantial unannotated scans are accessible. It is utilised in speech recognition \citep{deng2013machine} and bioinformatics \citep{tamposis2019semi}, where unlabelled data predominates. Graph-based and neural network-based semi-supervised approaches have demonstrated significant efficacy in managing intricate, high-dimensional data. However, specific drawbacks remain. The efficacy of these algorithms is significantly influenced by the quality of pseudo-labels and the soundness of the data assumptions. Discrepancies between the distributions of unlabelled and labelled data can lead to a decline in the model's accuracy. Recent study emphasises the advancement of more resilient and noise-resistant SSL algorithms to address these issues and optimize the utilisation of extensive unlabelled datasets.

\subsection{Self-Training Algorithm}

ST is a SSL technique in which a model trained on a limited labelled dataset assigns pseudo-labels to unlabelled samples with high confidence. The pseudo-labelled samples are subsequently incorporated into the training set, and the model undergoes recurrent retraining. It is straightforward, efficient, and extensively utilised in domains such as natural language processing \citep{du2021self}, computer vision \citep{ghiasi2021multi}, and speech recognition \citep{kahn2020self}. Various variants of ST encompass threshold-based \citep{scudder1965adaptive, tur2005combining}, proportion-based \citep{zou2018unsupervised}, curriculum learning-based \citep{zhang2021flexmatch}, majority-vote \citep{feofanov2024multi}, and adaptive thresholding methodologies \citep{wang2022freematch, amini2025self}. The threshold-based ST technique is the most used, as it focuses on picking only high-confidence predictions for pseudo-labelling to reduce mistakes and enhance model performance.

The threshold-based ST technique initiates by training a foundational classifier utilising the available labelled data. Upon completion of training, the classifier forecasts labels for the unlabelled samples and allocates confidence scores to each prediction. Only samples with confidence scores exceeding a designated threshold are selected and assigned pseudo-labels. The high-confidence pseudo-labelled samples are subsequently incorporated into the labelled dataset, and the model is retrained utilising this augmented dataset. The procedure is replicated throughout multiple rounds, with each subsequent model employing its enhanced decision boundary to classify further unlabelled data. The procedure persists until no additional unlabelled samples satisfy the confidence criterion or the model's performance stabilises. This iterative procedure enhances the model's generalisation capability while mitigating the risk of including wrong labels.

\subsection{Neurochaos Learning Algorithm}

The NL algorithm is a supervised learning model inspired by the brain that employs chaos theory for data analysis. Harikrishnan and Nithin Nagaraj originally proposed it for classification tasks in 2019 \citep{balakrishnan2019chaosnet}. The program replicates the chaotic firing of neurons in the human brain by the application of mathematical chaotic maps, including the Generalized Lüroth Series (GLS). Each input feature is transformed into a chaotic signal, from which pertinent characteristics such as firing time, firing rate, energy, and entropy are derived. These attributes are subsequently utilized by classifiers, like ChaosNet (based on cosine-similarity) or conventional machine learning techniques. Subsequent research expanded neural networks beyond classification to encompass regression tasks \citep{HENRY2025117213}, demonstrating their capability to predict continuous values. Consequently, NL has evolved into a comprehensive supervised learning technique capable of efficiently addressing both classification and regression tasks.

Multiple variations of NL have been created to enhance performance and adaptability. The principal variants include ChaosNet \citep{balakrishnan2019chaosnet}, ChaosFEX (NL) \citep{harikrishnan2020neurochaos}, Heterogeneous NL (HNL) \citep{as2023analysis}, Random Heterogeneous NL (RHNL) \citep{as7random}, Composition-based NL \citep{henry2025neurochaos}, and Augmented NL Regression Models \citep{HENRY2025117213}. These versions employ various combinations of chaotic maps, including the Logistic Map \citep{as2023analysis} and the Skew Tent Map \citep{sethi2023neurochaos}. NL and its variants have been employed in various applications, including forest fire prediction \citep{pant2025advancing}, protein classification \citep{anusree2024hypothetical, sneha2023biologically}, and medical data analysis \citep{harikrishnan2022classification}. The primary advantages of NL include its effective performance with less training data, retention of causal structures in the data \citep{nb2022causality}, interpretability, and reduced complexity compared to deep learning models.

\begin{figure}[h] 
\centering
\includegraphics[width=0.8\textwidth]{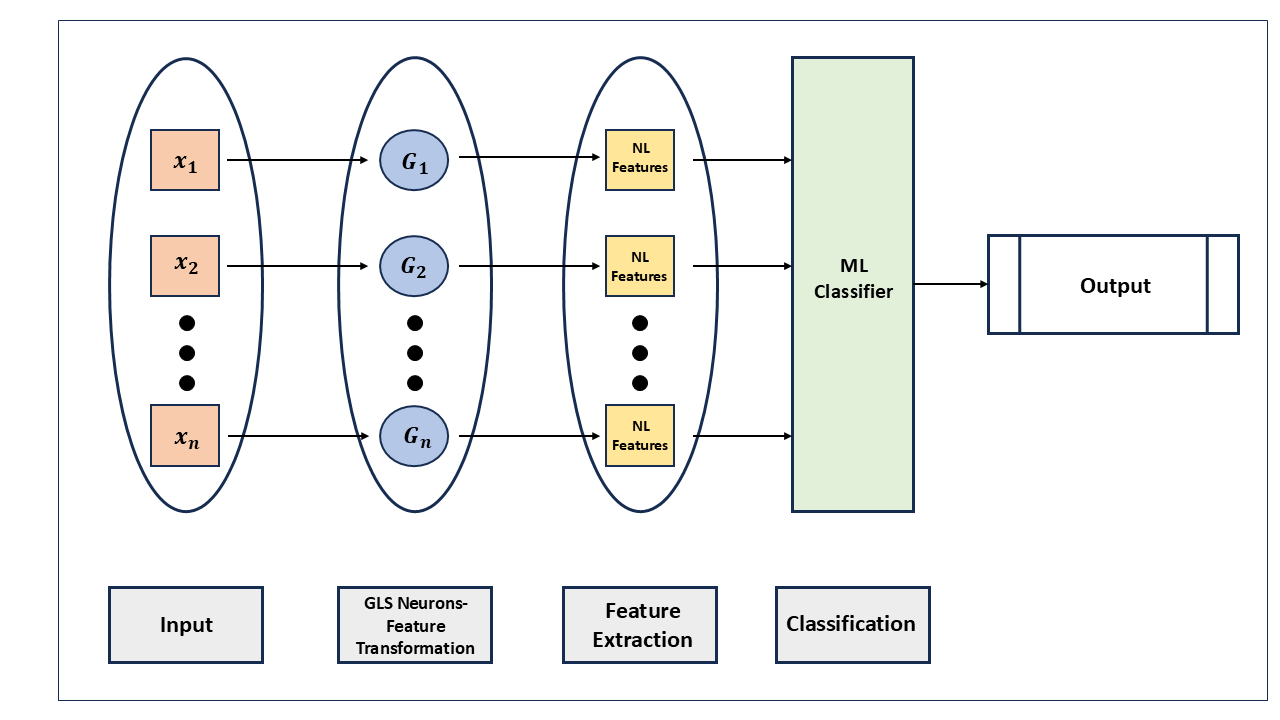}
\caption{NL Architecture}\label{nl_fig}

\end{figure}

The operational procedures of the NL architecture are straightforward and clearly delineated. The initial phase involves feature transformation, wherein each input value is processed by a chaotic neuron that exhibits chaotic firing until it attains the specified input value. The subsequent stage involves feature extraction, during which various attributes, including firing time, firing rate, energy, entropy, or the mean of the chaotic signal (termed Tracemean) \citep{henry2025simplified}, are computed from the neural trace. The third phase involves classification or regression, wherein the retrieved features are provided to a classifier like ChaosNet or to conventional machine learning classifiers or regression models. This approach enables NL to identify intricate nonlinear patterns within data. Figure \ref{nl_fig} illustrates the three phases of this algorithm: feature transformation, chaotic feature extraction, and classification.

\section{Proposed Methodology}
\label{sec:methodology}

This study evaluates the performance of the NL+ST framework. The aim of this hybrid NL+ST architecture is to enhance classification performance in scenarios with little labelled data or class imbalance in the dataset. The NL stage converts the original input features into chaos-based representations with skew-tent maps as the input layer neurons, whereas the ST stage expands the labelled set through confident predictions. This study utilises solely the firing-rate feature derived from NL as input for the ST classifier. The utilisation of the firing-rate characteristic guarantees a concise yet informative representation, enhancing classifier stability. This integrated method mitigates the prevalent issue of error propagation in ST and enhances class representation by using only high-confidence samples in each iteration. Five conventional machine learning techniques, specifically Random Forest (RF), AdaBoost (AB), Support Vector Machine (SVM), Logistic Regression (LR), and Gaussian Naive Bayes (GNB), are employed for evaluation. The outcomes are evaluated between standalone ST, standalone NL, and their combination (NL+ST). The macro F1-score serves as the performance statistic because to the significant imbalance in some datasets. Table~\ref{tab:datasets} presents the specifics of the ten benchmark datasets utilised for testing.

\begin{table}[htbp]
\centering
\caption{Details of the benchmark datasets used for evaluation.}
\label{tab:datasets}
\resizebox{\textwidth}{!}{%
\begin{tabular}{lccc}
\hline
\textbf{Dataset} & \textbf{Number of Samples} & \textbf{Number of Classes} & \textbf{Class Proportion Ratio} \\
\hline
\textit{Iris} & 150 & 3 & 1.0:1.0:1.0 \\
\textit{Wine} & 178 & 3 & 1.2:1.5:1.0 \\
\textit{Breast Cancer Wisconsin} & 569 & 2 & 1.0:1.7 \\
\textit{Haberman’s Survival} & 306 & 2 & 2.8:1.0 \\
\textit{Ionosphere} & 351 & 2 & 1.0:1.8 \\
\textit{Statlog (Heart)} & 270 & 2 & 1.2:1.0 \\
\textit{Seeds} & 210 & 3 & 1.0:1.0:1.0 \\
\textit{Palmer Penguins Dataset} & 333 & 3 & 2.1:1.0:1.8 \\
\textit{Pima Indians Diabetes Dataset} & 768 & 2 & 1.9:1.0 \\
\textit{Glass Identification} & 214 & 6 & 7.8:8.4:1.4:3.2:1.0:1.9 \\
\hline
\end{tabular}%
}
\end{table}

The NL hyperparameters are fixed as follows: discrimination threshold $b = 0.499$ as it gives maximum entropy to the symbolic sequence \citep{henry2025simplified} and neighbourhood width $\epsilon = 0.25$. The ST confidence level is established at 75\%. The initial neural activity $q$ is calibrated within the interval $[0,1]$ by a five-fold cross-validation conducted independently for each dataset and classifier. The output of the NL feature extraction block is the firing rate vector for each data sample. For a specified input feature $x_{ij}$, the neuron commences with an initial activity $q$ and iterates the chaotic map until its trajectory enters a $\epsilon$-neighbourhood of $x_{ij}$. A binary symbolic sequence is subsequently created utilising $b = 0.499$. The firing rate for this characteristic is calculated as the proportion of trajectory points above the discrimination threshold. The rates constitute the modified feature vector that acts as the input for the ST classifier. In all studies, 85\% of the training data is designated as unlabelled samples, whereas only 15\% is utilised as labelled data for the training of the ST classifier. 

\begin{figure}[H] 
\centering
\includegraphics[width=0.8\textwidth]{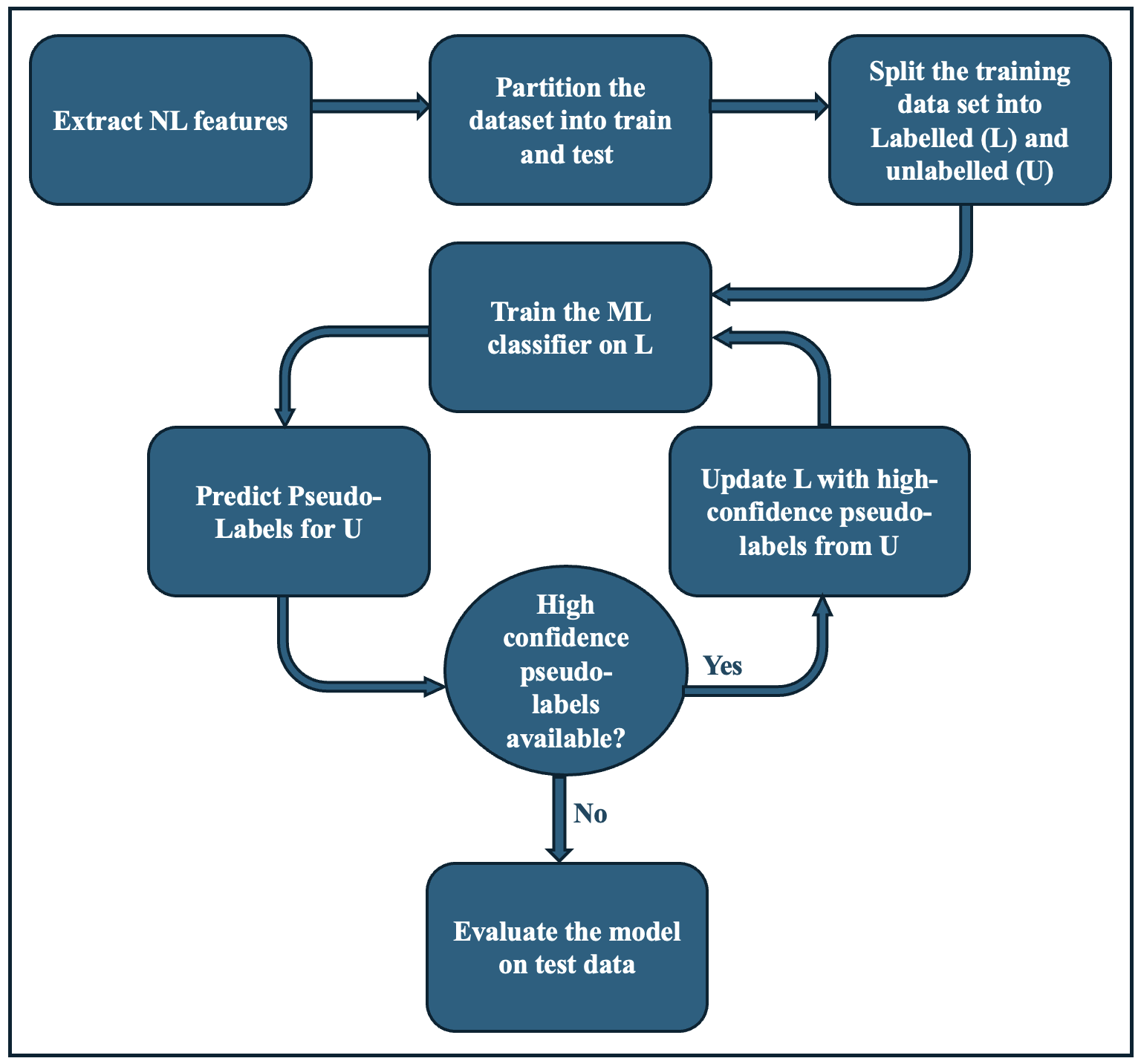}
\caption{Flowchart illustrating the steps involved in the proposed NL+ST methodology.}
\label{fig:flowchart}
\end{figure}

The comprehensive methodology comprises multiple sequential steps. Initially, all input features within each dataset are standardised to the interval [0,1]. Firing-rate features of NL are derived from all samples utilising constant values of $b$ and $\epsilon$. The 80\% training data from each dataset is subsequently partitioned into a labelled set ($L$) and an unlabelled set ($U$). The base classifier is trained on the labelled set ($L$) utilising these firing-rate vectors. The trained classifier subsequently predicts pseudo-labels for the unlabelled set ($U$), selecting examples with prediction confidence of 75\% or higher. The high-confidence samples are incorporated into the labelled set ($L$), and the classifier is then retrained. The selection and retraining procedure continues until no further samples satisfy the threshold criteria. The trained model is ultimately assessed on the 20\% test partition, and the resulting F1-scores are documented. The comprehensive workflow of the proposed NL+ST methodology is depicted in Figure~\ref{fig:flowchart}.

\section{Results and Discussion}
\label{sec:results}

The optimum values of $q$ for each dataset and classifier are shown in the relevant tables within their respective subsections. All comparative result tables and bar graphs illustrating the performance of the three models on each dataset are included in the same subsections for clarity.


\subsection{Random Forest}

\begin{figure}[H]
\centering
\includegraphics[width=0.8\textwidth]{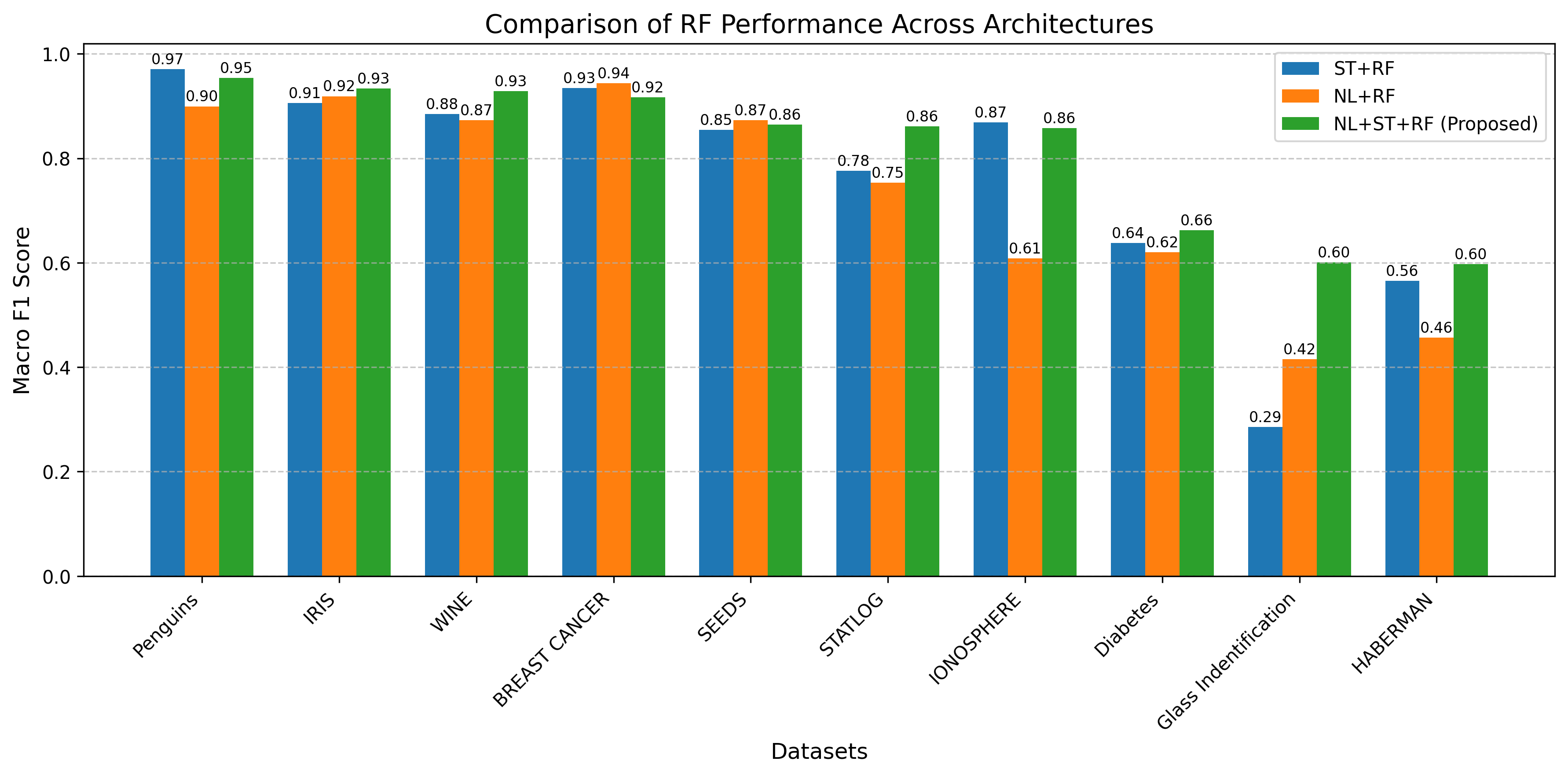}
\caption{Bar graph comparison of Random Forest performance across different architectures.}
\label{fig:rf_bar}
\end{figure}

The macro F1 scores of the RF classifier over all 10 datasets are presented in Table~\ref{tab:f1_rf}. The results indicate that the suggested NL+ST model (NL+ST+RF) outperforms both the standalone ST RF (ST+RF) and the standalone NL RF (NL+RF) across the majority of datasets. The enhancement is particularly evident in datasets like \textit{Glass Identification}, \textit{Statlog (Heart)}, and \textit{Haberman’s Survival}, where the proposed technique attains superior F1 scores. These datasets exhibit either a scarcity of labelled samples or a significant class imbalance, wherein the integrated design leverages the noise-resistant feature extraction of the NL stage and the confidence-driven selection of the ST stage. In several instances, such as \textit{Breast Cancer Wisconsin} and \textit{Iris}, the scores are analogous across all three models, suggesting that the baseline classifier already exhibits strong performance on balanced datasets with adequate samples. The findings indicate that the integration of ST with NL improves classification robustness and generalisation in scenarios with limited labelled data. The tuned hyperparameter values ($q$) for the RF classifier, derived from 5-fold cross-validation, are also presented in Table~\ref{tab:f1_rf}. Figure~\ref{fig:rf_bar} illustrates the bar graph comparing the macro F1 scores of the RF classifier across the three architectures.

\begin{table}[htbp]
\centering
\caption{Tuned $q$ values and macro F1 scores of RF classifier using different architectures:ST+RF, NL+RF, and the proposed NL+ST+RF.}
\label{tab:f1_rf}
\resizebox{\textwidth}{!}{%
\begin{tabular}{lcccc}
\hline
\textbf{Dataset} & \textbf{$q$ Value} & \textbf{ST+RF} & \textbf{NL+RF} & \textbf{NL+ST+RF} \\
\hline
\textit{Iris} & 0.956 & 0.9057 & 0.9188 & \textbf{0.9333} \\
\textit{Wine} & 0.445 & 0.8843 & 0.873 & \textbf{0.9285} \\
\textit{Breast Cancer Wisconsin} & 0.996 & 0.9344 & \textbf{0.9435} & 0.9164 \\
\textit{Haberman’s Survival} & 0.708 & 0.5649 & 0.4564 & \textbf{0.5974} \\
\textit{Ionosphere} & 0.821 & \textbf{0.869} & 0.6081 & 0.858 \\
\textit{Statlog (Heart)} & 0.986 & 0.7757 & 0.7531 & \textbf{0.8611} \\
\textit{Seeds} & 0.753 & 0.8543 & \textbf{0.8725} & 0.8645 \\
\textit{Palmer Penguins Dataset} & 0.967 & \textbf{0.9708} & 0.8994 & 0.9538 \\
\textit{Pima Indians Diabetes Dataset} & 0.916 & 0.6375 & 0.6198 & \textbf{0.662} \\
\textit{Glass Identification} & 0.942 & 0.2853 & 0.4155 & \textbf{0.6005} \\
\hline
\end{tabular}%
}
\end{table}

\subsection{AdaBoost}

\begin{figure}[H]
\centering
\includegraphics[width=0.8\textwidth]{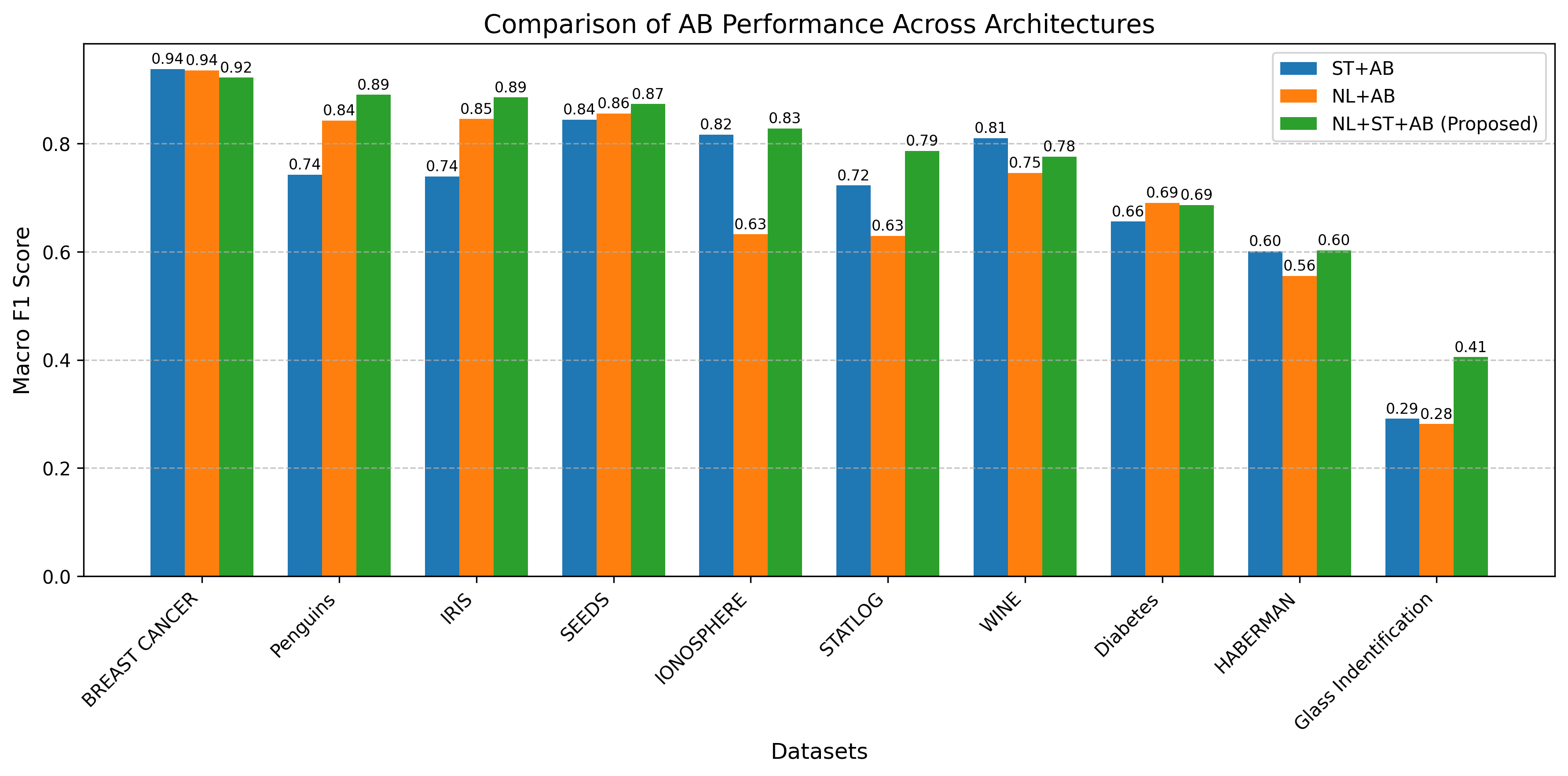}
\caption{Bar graph comparison of AdaBoost performance across different architectures.}
\label{fig:ab_bar}
\end{figure}

The macro F1 scores and the optimised $q$ values for AB are presented in Table~\ref{tab:f1_ab}. The proposed NL+ST+AB model regularly attains superior F1 scores compared to the individual versions. AB advantages from the improved separability of the NL features, which offer more distinctive input to the weak learners. The ST step enhances the ensemble by integrating high-confidence pseudo-labelled samples. The most significant performance improvement is noted in the \textit{Glass Identification} dataset, which exhibits greater imbalance. Figure~\ref{fig:ab_bar} illustrates the comparative performance of the AB classifier across the ST+AB, NL+AB, and the proposed NL+ST+AB architectures.

\begin{table}[htbp]
\centering
\caption{Tuned $q$ values and macro F1 scores of AB classifier using different architectures: ST+AB, NL+AB, and the proposed NL+ST+AB.}
\label{tab:f1_ab}
\resizebox{\textwidth}{!}{%
\begin{tabular}{lcccc}
\hline
\textbf{Dataset} & \textbf{$q$ Value} & \textbf{ST+AB} & \textbf{NL+AB} & \textbf{NL+ST+AB} \\
\hline
\textit{Iris} & 0.973 & 0.7396 & 0.8459 & \textbf{0.8857} \\
\textit{Wine} & 0.983 & \textbf{0.8105} & 0.7458 & 0.7763 \\
\textit{Breast Cancer Wisconsin} & 0.994 & \textbf{0.9381} & 0.9357 & 0.9222 \\
\textit{Haberman’s Survival} & 0.708 & 0.6012 & 0.5554 & \textbf{0.6026} \\
\textit{Ionosphere} & 0.831 & 0.8168 & 0.6323 & \textbf{0.8281} \\
\textit{Statlog (Heart)} & 0.58 & 0.7229 & 0.6296 & \textbf{0.7869} \\
\textit{Seeds} & 0.767 & 0.8447 & 0.8559 & \textbf{0.8734} \\
\textit{Palmer Penguins Dataset} & 0.967 & 0.7428 & 0.843 & \textbf{0.8906} \\
\textit{Pima Indians Diabetes Dataset} & 0.948 & 0.6561 & \textbf{0.6905} & 0.6865 \\
\textit{Glass Identification} & 0.58 & 0.2911 & 0.2814 & \textbf{0.405} \\
\hline
\end{tabular}%
}
\end{table}

\subsection{Support Vector Machine}

\begin{figure}[H]
\centering
\includegraphics[width=0.8\textwidth]{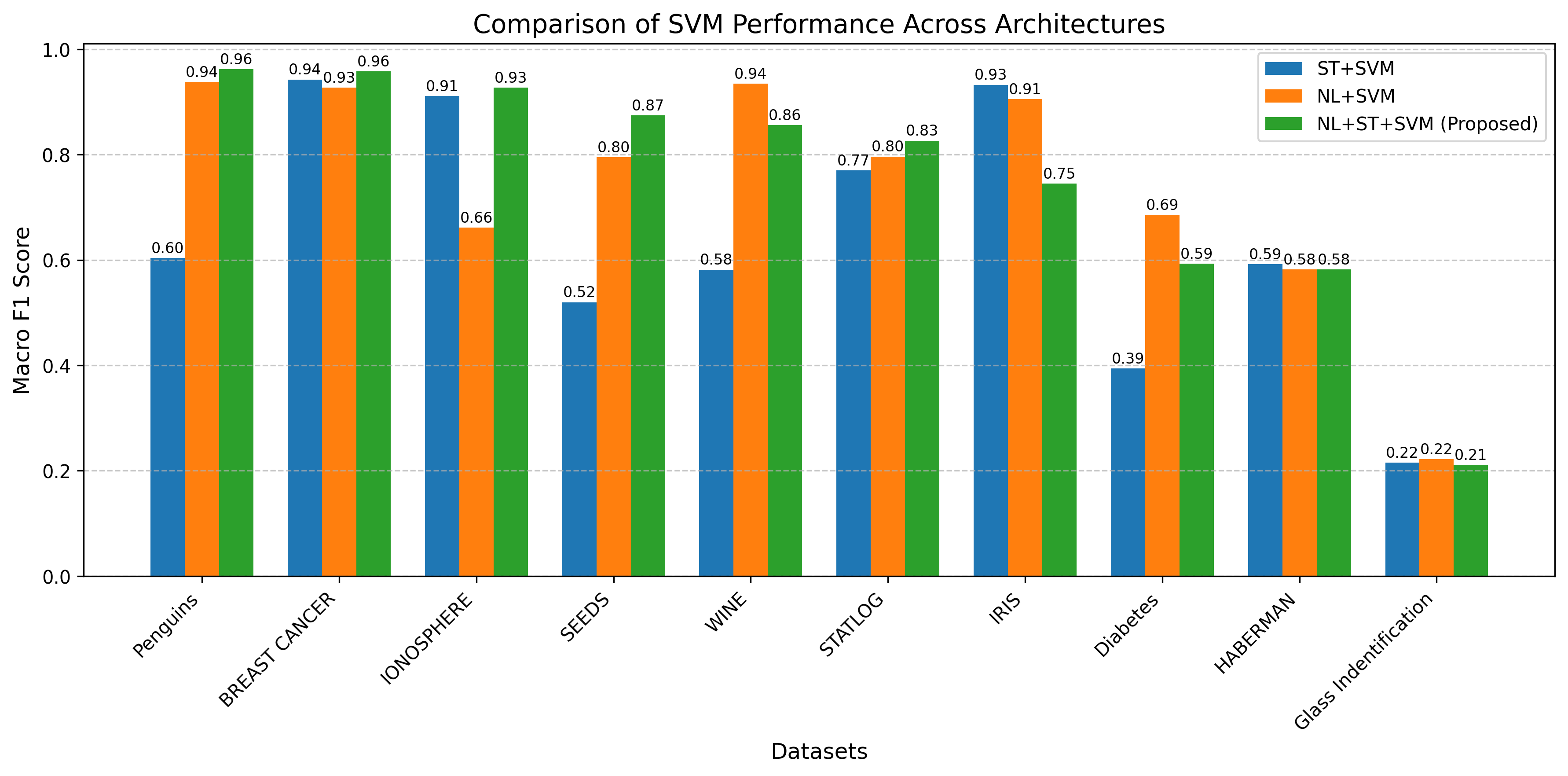}
\caption{Bar graph comparison of Support Vector Machine performance across different architectures.}
\label{fig:svm_bar}
\end{figure}

Table~\ref{tab:f1_svm} displays the F1 scores for the SVM classifier, and the associated optimised $q$ values. The suggested NL+ST+SVM yields superior results for fifty percent of the datasets examined, suggesting that chaos-induced features augment class differentiation in the feature space. The kernel-based decision boundary of SVM is enhanced by the nonlinear transformations generated during the NL stage. ST enhances the model's adaptability when labelled data is scarce. For datasets such as \textit{Breast Cancer Wisconsin} and \textit{Statlog (Heart)}, the improvement is moderate yet constant, whereas more substantial advances are noted for the \textit{Seeds} dataset. Figure~\ref{fig:svm_bar} depicts the bar graph that compares the macro F1 scores achieved by the SVM classifier across all three architectures.

\begin{table}[htbp]
\centering
\caption{Tuned $q$ values and macro F1 scores of SVM classifier using different architectures: ST+SVM, NL+SVM, and the proposed NL+ST+SVM.}
\label{tab:f1_svm}
\resizebox{\textwidth}{!}{%
\begin{tabular}{lcccc}
\hline
\textbf{Dataset} & \textbf{$q$ Value} & \textbf{ST+SVM} & \textbf{NL+SVM} & \textbf{NL+ST+SVM} \\
\hline
\textit{Iris} & 0.947 & \textbf{0.9327} & 0.9056 & 0.7453 \\
\textit{Wine} & 0.431 & 0.5816 & \textbf{0.935} & 0.856 \\
\textit{Breast Cancer Wisconsin} & 0.988 & 0.9426 & 0.9273 & \textbf{0.9584} \\
\textit{Haberman’s Survival} & 0.345 & \textbf{0.5922} & 0.5823 & 0.5823 \\
\textit{Ionosphere} & 0.959 & 0.9115 & 0.6616 & \textbf{0.927} \\
\textit{Statlog (Heart)} & 0.996 & 0.7699 & 0.7964 & \textbf{0.8265} \\
\textit{Seeds} & 0.951 & 0.5199 & 0.795 & \textbf{0.8743}\\
\textit{Palmer Penguins Dataset} & 0.719 & 0.6037 & 0.938 & \textbf{0.9625} \\
\textit{Pima Indians Diabetes Dataset} & 0.923 & 0.3943 & \textbf{0.686} & 0.5928 \\
\textit{Glass Identification} & 0.997 & 0.2151 & \textbf{0.2222} & 0.211 \\
\hline
\end{tabular}%
}
\end{table}

\subsection{Logistic Regression}

\begin{figure}[H]
\centering
\includegraphics[width=0.8\textwidth]{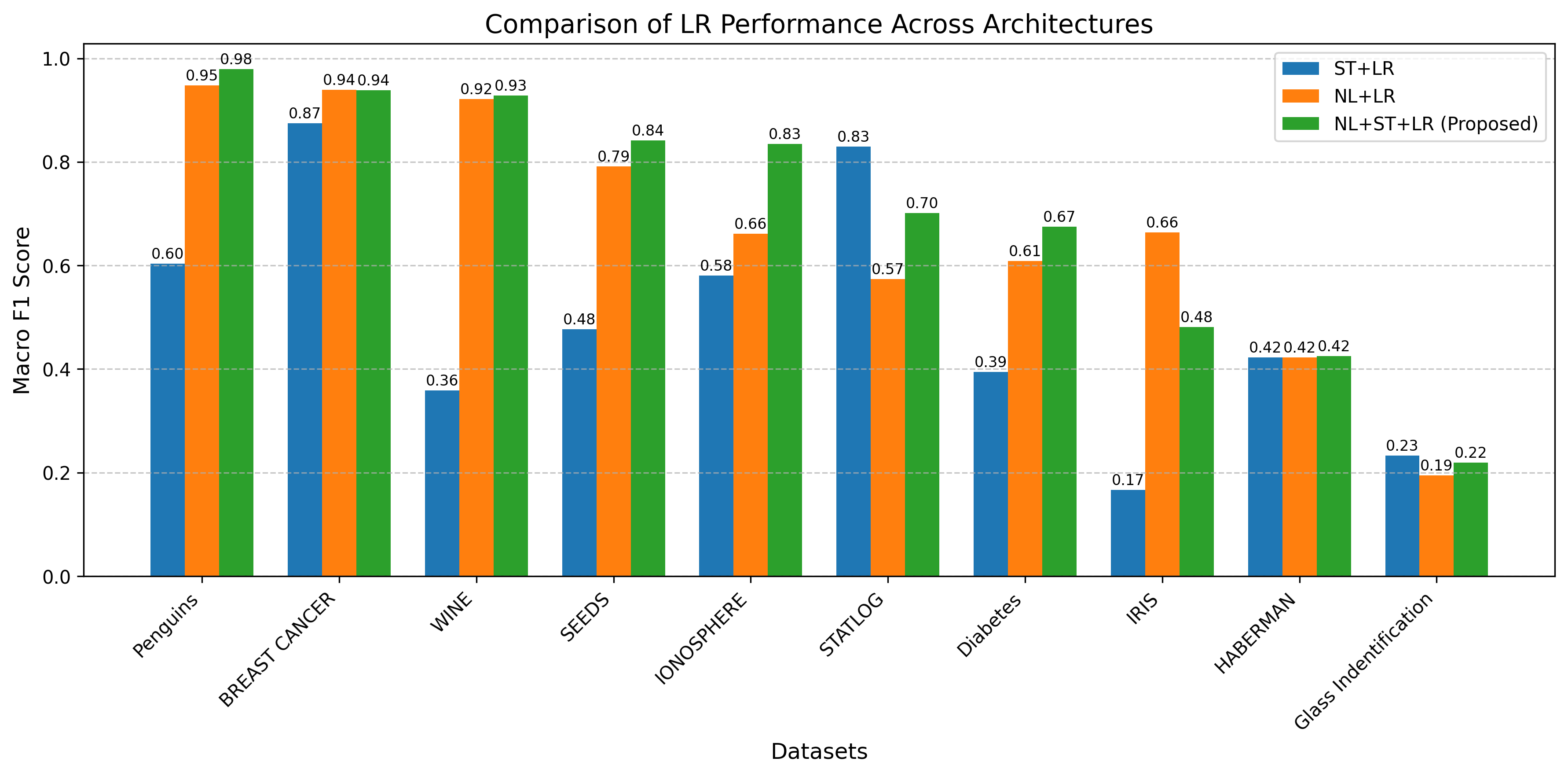}
\caption{Bar graph comparison of Logistic Regression performance across different architectures.}
\label{fig:lr_bar}
\end{figure}

The macro F1 scores and the optimised $q$ values for the LR classifier are presented in Table~\ref{tab:f1_lr}. The proposed NL+ST+LR model outperforms the individual components, particularly on datasets characterised by pronounced nonlinear class boundaries, such as the \textit{Ionosphere} and \textit{Pima Indians Diabetes Dataset} datasets. Despite LR being a linear model, the chaos-based feature extraction incorporates nonlinear dynamics into the input space, enhancing separability. The ST component enhances robustness and enables the model to utilise supplementary confident samples, resulting in elevated average F1 scores. Figure~\ref{fig:lr_bar} illustrates the comparison of LR classifier outcomes utilising the ST+LR, NL+LR, and the proposed NL+ST+LR frameworks.

\begin{table}[htbp]
\centering
\caption{Tuned $q$ values and macro F1 scores of LR classifier using different architectures: ST+LR, NL+LR, and the proposed NL+ST+LR.}
\label{tab:f1_lr}
\resizebox{\textwidth}{!}{
\begin{tabular}{lcccc}
\hline
\textbf{Dataset} & \textbf{$q$ Value} & \textbf{ST+LR} & \textbf{NL+LR} & \textbf{NL+ST+LR} \\
\hline
\textit{Iris} & 0.5 & 0.1667 & \textbf{0.6641} & 0.4812 \\
\textit{Wine} & 0.936 & 0.3588 & 0.9215 & \textbf{0.9278} \\
\textit{Breast Cancer Wisconsin} & 0.982 & 0.8745 & \textbf{0.9389} & 0.9383 \\
\textit{Haberman’s Survival} & 0.001 & 0.4226 & 0.4226 & \textbf{0.425} \\
\textit{Ionosphere} & 0.933 & 0.5805 & 0.6615 & \textbf{0.8344} \\
\textit{Statlog (Heart)} & 0.571 & \textbf{0.8296} & 0.5736 & 0.7013 \\
\textit{Seeds} & 0.988 & 0.477 & 0.791 & \textbf{0.8411} \\
\textit{Palmer Penguins Dataset} & 0.719 & 0.6037 & 0.9478 & \textbf{0.9793} \\
\textit{Pima Indians Diabetes Dataset} & 0.521 & 0.3943 & 0.6089 & \textbf{0.6748} \\
\textit{Glass Identification} & 0.628 & \textbf{0.2327} & 0.1945 & 0.2194 \\
\hline
\end{tabular}%
}
\end{table}

\subsection{Gaussian Naive Bayes}

\begin{figure}[H]
\centering
\includegraphics[width=0.8\textwidth]{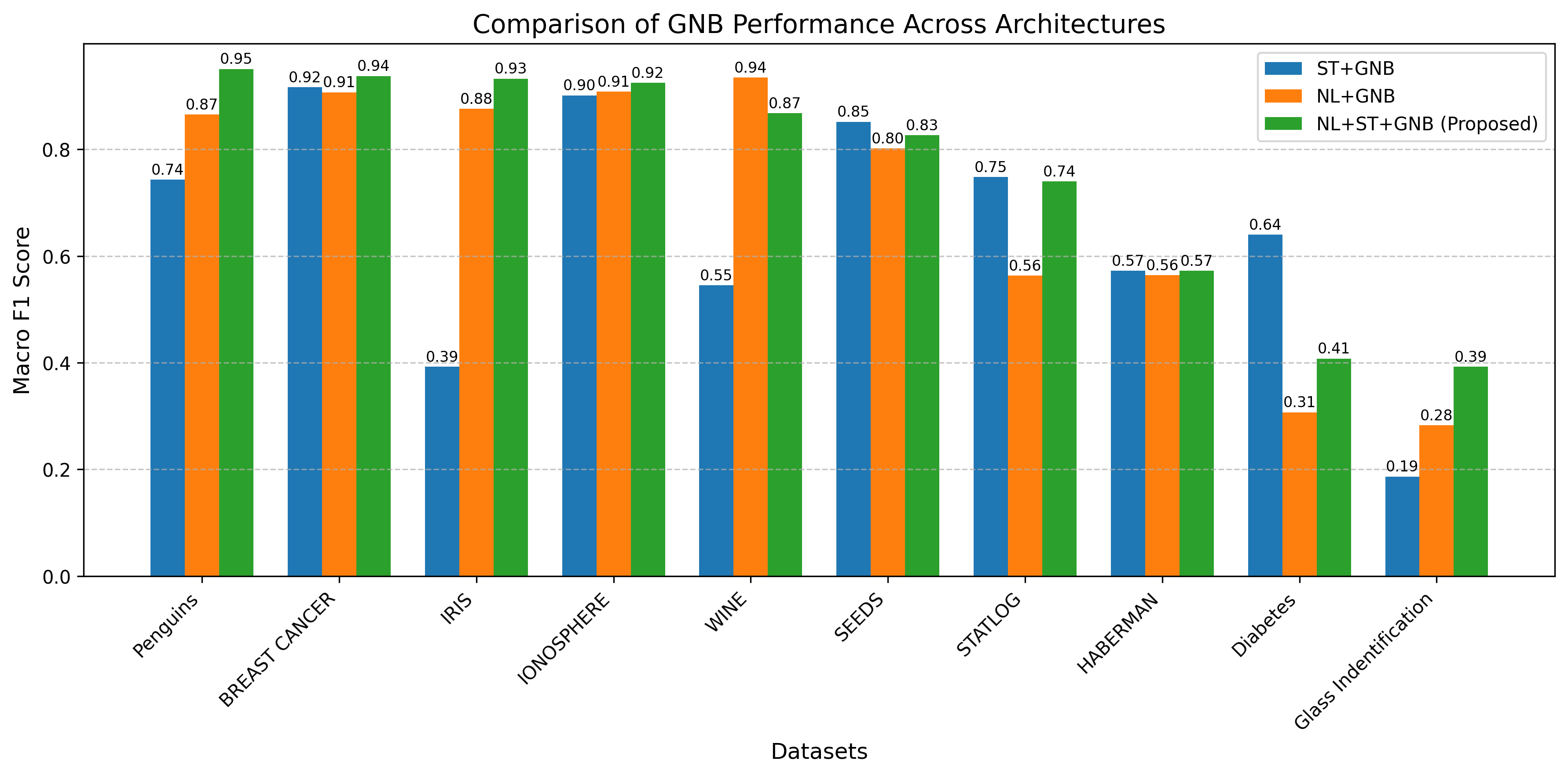}
\caption{Bar graph comparison of Gaussian Naive Bayes performance across different architectures.}
\label{fig:gnb_bar}
\end{figure}

The outcomes and the optimised $q$ values for GNB are presented in Table~\ref{tab:f1_gnb}. The NL+ST+GNB model attains equivalent F1 scores across the majority of datasets. The probabilistic characteristics of Naïve Bayes are enhanced by the stable and noise-resistant features produced during the NL stage. In scenarios characterised by limited sample sizes and skewed data, such as the \textit{Glass Identification} dataset, the integrated model demonstrates markedly superior performance compared to the individual algorithms. The results indicate that even basic models can benefit from the suggested feature transformation and semi-supervised framework. Figure~\ref{fig:gnb_bar} presents the bar graph comparing the macro F1 scores for the GNB classifier across the ST+GNB, NL+GNB, and suggested NL+ST+GNB architectures.

\begin{table}[htbp]
\centering
\caption{Tuned $q$ values and macro F1 scores of GNB classifier using different architectures: ST+GNB, NL+GNB, and the proposed NL+ST+GNB.}
\label{tab:f1_gnb}
\resizebox{\textwidth}{!}{%
\begin{tabular}{lcccc}
\hline
\textbf{Dataset} & \textbf{$q$ Value} & \textbf{ST+GNB} & \textbf{NL+GNB} & \textbf{NL+ST+GNB} \\
\hline
\textit{Iris} & 0.957 & 0.3926 & 0.876 & \textbf{0.9327} \\
\textit{Wine} & 0.437 & 0.5451 & \textbf{0.935} & 0.8679 \\
\textit{Breast Cancer Wisconsin} & 0.588 & 0.9164 & 0.907 & \textbf{0.9376} \\
\textit{Haberman’s Survival} & 0.808 & 0.5722 & 0.5641 & 0.5722 \\
\textit{Ionosphere} & 0.76 & 0.9013 & 0.9083 & \textbf{0.9248} \\
\textit{Statlog (Heart)} & 0.91 & \textbf{0.748} & 0.5634 & 0.7398 \\
\textit{Seeds} & 0.76 & \textbf{0.8518} & 0.8017 & 0.8267 \\
\textit{Palmer Penguins Dataset} & 0.955 & 0.7433 & 0.8652 & \textbf{0.9507} \\
\textit{Pima Indians Diabetes Dataset} & 0.87 & \textbf{0.6403} & 0.3072 & 0.4078 \\
\textit{Glass Identification} & 0.997 & 0.1867 & 0.2825 & \textbf{0.3927} \\
\hline
\end{tabular}%
}
\end{table}

\subsection{Comparative Analysis among Classifiers}

The performance gain values are estimated to find the percentage of improvement achieved by the proposed hybrid NL+ST architecture over the standalone ST approach for each classifier and dataset. It is computed using the expression
\[
\text{Gain}(\%) = \frac{\text{F1}_{\text{NL+ST}} - \text{F1}_{\text{ST}}}{\text{F1}_{\text{ST}}} \times 100,
\]
which normalizes the improvement relative to the baseline ST performance. This formulation provides a consistent and fair comparison of performance enhancement across different machine-learning models and datasets, even when the absolute F1-scores vary significantly. A positive gain value indicates that the hybrid NL+ST strategy improves the classification accuracy compared to the standalone ST method. Table~\ref{tab:gain_values} summarises the improvement achieved in the performance of classifiers on various datasets.

The table shows clear differences in how each algorithm benefits from the hybrid NL+ST architecture. Among the five classifiers, LR achieves the highest overall improvement, with large positive gains across many datasets. LR shows very high gains for the \textit{Iris}, \textit{Wine}, \textit{Seeds}, \textit{Pima Indians Diabetes Dataset}, \textit{Palmer Penguins Dataset}, and \textit{Ionosphere} datasets. This indicates that LR benefits strongly from the nonlinear firing-rate features produced by NL. Since LR is a linear classifier, the NL transformation likely makes the classes more separable, allowing LR to learn better decision boundaries even with limited labelled data. SVM also records high improvements in several datasets, showing that margin-based classifiers can effectively use the richer nonlinear structure introduced by NL. RF and AB show moderate to high gains in a few datasets, while GNB has mixed performance with both positive and negative gains.

\begin{figure}[H]
\centering
\includegraphics[width=0.8\textwidth]{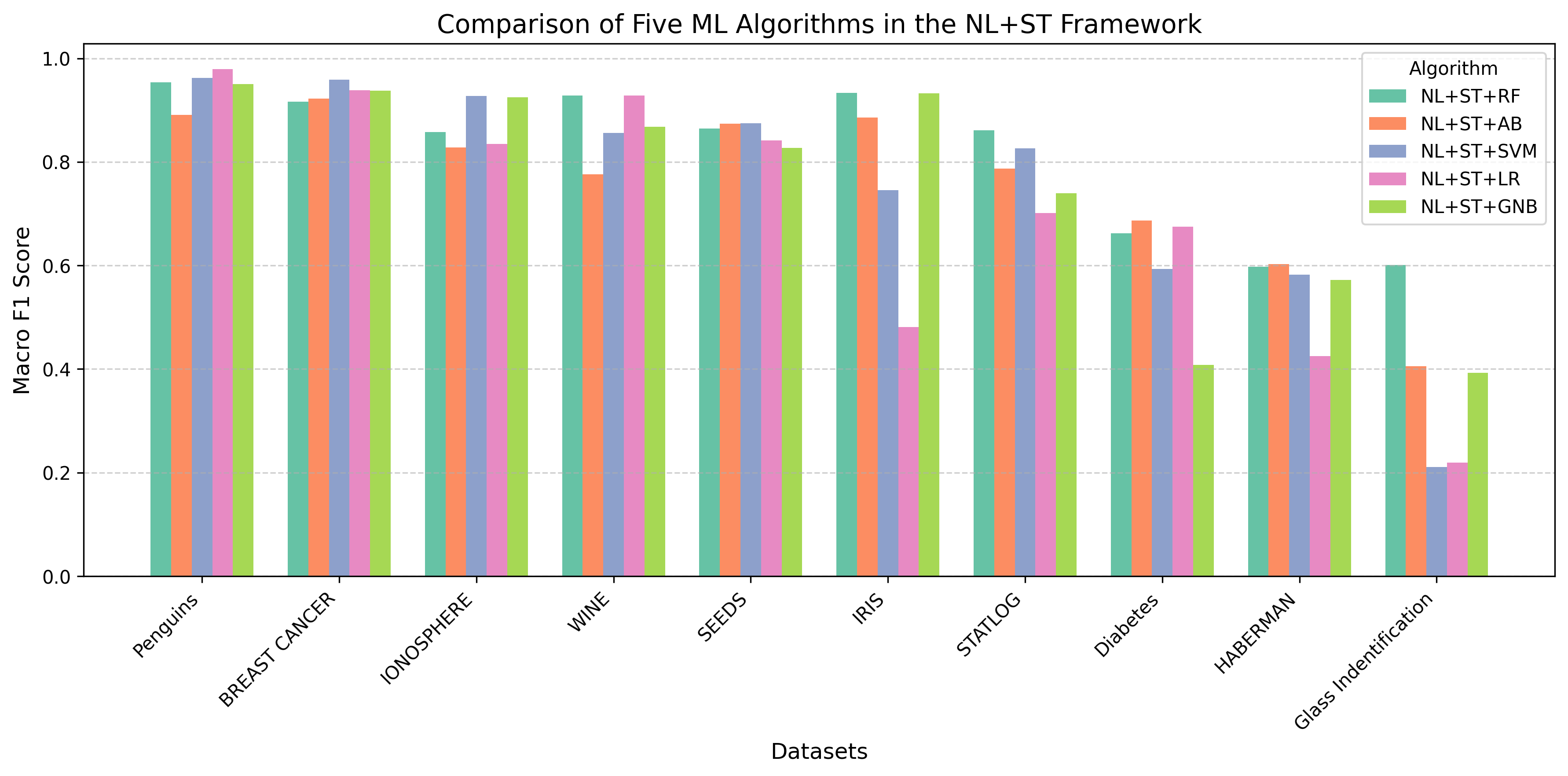}
\caption{Comparison of five ML algorithms in the NL+ST framework across ten benchmark datasets.}
\label{fig:algo_comparison}
\end{figure}

Among all datasets, the \textit{Iris} dataset shows the highest overall improvement across algorithms, especially under LR(188.66) and GNB(137.57). This suggests that the NL feature transformation enhances class separability in this dataset more effectively than in others. The \textit{Glass Identification} dataset also benefits significantly for algorithms such as RF(110.48) and GNB(110.34). Figure~\ref{fig:algo_comparison} illustrates the performance comparison of five machine learning algorithms within the NL+ST framework across ten benchmark datasets. These results show that the hybrid NL+ST method is most effective when applied to datasets where the NL transformation can strengthen class boundary separation and when used with classifiers that rely on improved feature separability.

\begin{table}[H]
\centering
\caption{Performance gain (\%) on NL+ST model over the standalone ST.}
\label{tab:gain_values}
\begin{tabular}{clr}
\hline
\textbf{Algorithm} & \textbf{Dataset} & \textbf{Gain (\%)} \\
\hline
\multirow{8}{*}{\centering\textbf{LR}} 
& {\coltwofont \textit{Iris}} & {\colthreefont 188.66} \\
& {\coltwofont \textit{Wine}} & {\colthreefont 158.58} \\
& {\coltwofont \textit{Seeds}} & {\colthreefont 76.33} \\
& {\coltwofont \textit{Pima Indians Diabetes Dataset}} & {\colthreefont 71.14} \\
& {\coltwofont \textit{Palmer Penguins Dataset}} & {\colthreefont 62.22} \\
& {\coltwofont \textit{Ionosphere}} & {\colthreefont 43.74} \\
& {\coltwofont \textit{Breast Cancer Wisconsin}} & {\colthreefont 7.30} \\
& {\coltwofont \textit{Haberman’s Survival}} & {\colthreefont 0.57} \\
\hline
\multirow{7}{*}{\centering\textbf{RF}} 
& {\coltwofont \textit{Glass Identification}} & {\colthreefont 110.48} \\
& {\coltwofont \textit{Statlog (Heart)}} & {\colthreefont 11.01} \\
& {\coltwofont \textit{Haberman’s Survival}} & {\colthreefont 5.75} \\
& {\coltwofont \textit{Wine}} & {\colthreefont 5.00} \\
& {\coltwofont \textit{Pima Indians Diabetes Dataset}} & {\colthreefont 3.84} \\
& {\coltwofont \textit{Iris}} & {\colthreefont 3.05} \\
& {\coltwofont \textit{Seeds}} & {\colthreefont 1.19} \\
\hline
\multirow{6}{*}{\centering\textbf{GNB}} 
& {\coltwofont \textit{Iris}} & {\colthreefont 137.57} \\
& {\coltwofont \textit{Glass Identification}} & {\colthreefont 110.34} \\
& {\coltwofont \textit{Wine}} & {\colthreefont 59.22} \\
& {\coltwofont \textit{Palmer Penguins Dataset}} & {\colthreefont 27.90} \\
& {\coltwofont \textit{Ionosphere}} & {\colthreefont 2.61} \\
& {\coltwofont \textit{Breast Cancer Wisconsin}} & {\colthreefont 2.31} \\
\hline
\multirow{7}{*}{\centering\textbf{SVM}} 
& {\coltwofont \textit{Seeds}} & {\colthreefont 68.17} \\
& {\coltwofont \textit{Palmer Penguins Dataset}} & {\colthreefont 59.43} \\
& {\coltwofont \textit{Pima Indians Diabetes Dataset}} & {\colthreefont 50.34} \\
& {\coltwofont \textit{Wine}} & {\colthreefont 47.18} \\
& {\coltwofont \textit{Statlog (Heart)}} & {\colthreefont 7.35} \\
& {\coltwofont \textit{Ionosphere}} & {\colthreefont 1.70} \\
& {\coltwofont \textit{Breast Cancer Wisconsin}} & {\colthreefont 1.68} \\
\hline
\multirow{8}{*}{\centering\textbf{AB}} 
& {\coltwofont \textit{Glass Identification}} & {\colthreefont 39.13} \\
& {\coltwofont \textit{Palmer Penguins Dataset}} & {\colthreefont 19.90} \\
& {\coltwofont \textit{Iris}} & {\colthreefont 19.75} \\
& {\coltwofont \textit{Statlog (Heart)}} & {\colthreefont 8.85} \\
& {\coltwofont \textit{Pima Indians Diabetes Dataset}} & {\colthreefont 4.63} \\
& {\coltwofont \textit{Seeds}} & {\colthreefont 3.40} \\
& {\coltwofont \textit{Ionosphere}} & {\colthreefont 1.38} \\
& {\coltwofont \textit{Haberman’s Survival}} & {\colthreefont 0.23} \\
\hline
\end{tabular}
\end{table}

\section{Conclusion}
\label{sec:conclusion}

This study suggested a hybrid SSL framework that combines the neural network architecture with a threshold-based ST method. The model employs a chaos-inspired firing-rate feature to encapsulate nonlinear dynamics in the data and utilises ST to efficiently exploit unlabelled samples. Experiments were performed utilising ten benchmark datasets and five machine learning classifiers. The findings indicate that the suggested NL+ST architecture regularly attains superior macro F1 scores relative to the independent NL and ST models. The enhancement is particularly significant for datasets characterised by limited and imbalanced distribution, such as \textit{Iris}(188.66) and \textit{Glass Identification}(110.48). RF and LR classifiers exhibit the most consistent and substantial performance improvements.

The findings validate that the integration of chaos-based feature transformation and confidence-based ST improves classification efficacy. The suggested methodology provides a comprehensive and adaptable framework that may integrate several classifiers and semi-supervised methods. Future research will investigate the expansion of this hybrid architecture for unsupervised learning tasks, adaptive threshold selection, and integration with deep learning frameworks. This method possesses significant promise for utilisation in biological and environmental fields where labelled data is limited.

\bibliographystyle{elsarticle-harv}
\bibliography{SelfTraining_NL}

@Article{jin2025advancements,
  author    = {Jin, Chengcheng and Ng, Theam Foo and Ibrahim, Haidi},
  journal   = {AI},
  title     = {Advancements in semi-supervised deep learning for brain tumor segmentation in mri: A literature review},
  year      = {2025},
  number    = {7},
  pages     = {153},
  volume    = {6},
  publisher = {MDPI},
}

@Article{deng2013machine,
  author    = {Deng, Li and Li, Xiao},
  journal   = {IEEE Transactions on Audio, Speech, and Language Processing},
  title     = {Machine learning paradigms for speech recognition: An overview},
  year      = {2013},
  number    = {5},
  pages     = {1060--1089},
  volume    = {21},
  publisher = {IEEE},
}

@Article{duarte2023review,
  author    = {Duarte, Jos{\'e} Marcio and Berton, Lilian},
  journal   = {Artificial intelligence review},
  title     = {A review of semi-supervised learning for text classification},
  year      = {2023},
  number    = {9},
  pages     = {9401--9469},
  volume    = {56},
  publisher = {Springer},
}

@Article{silva2016survey,
  author    = {Silva, Nadia Felix F Da and Coletta, Luiz FS and Hruschka, Eduardo R},
  journal   = {ACM Computing Surveys (CSUR)},
  title     = {A survey and comparative study of tweet sentiment analysis via semi-supervised learning},
  year      = {2016},
  number    = {1},
  pages     = {1--26},
  volume    = {49},
  publisher = {ACM New York, NY, USA},
}

@InProceedings{blum1998combining,
  author       = {Blum, Avrim and Mitchell, Tom},
  booktitle    = {Proceedings of the eleventh annual conference on Computational learning theory},
  title        = {Combining labeled and unlabeled data with co-training},
  year         = {1998},
  month        = {July},
  organization = {ACM},
  pages        = {92--100},
}

@Article{van2020survey,
  author    = {Van Engelen, Jesper E and Hoos, Holger H},
  journal   = {Machine learning},
  title     = {A survey on semi-supervised learning},
  year      = {2020},
  number    = {2},
  pages     = {373--440},
  volume    = {109},
  publisher = {Springer},
}

@Article{amini2025self,
  author    = {Amini, Massih-Reza and Feofanov, Vasilii and Pauletto, Loic and Hadjadj, Lies and Devijver, Emilie and Maximov, Yury},
  journal   = {Neurocomputing},
  title     = {Self-training: A survey},
  year      = {2025},
  pages     = {128904},
  volume    = {616},
  publisher = {Elsevier},
}

@Article{song2022graph,
  author    = {Song, Zixing and Yang, Xiangli and Xu, Zenglin and King, Irwin},
  journal   = {IEEE Transactions on Neural Networks and Learning Systems},
  title     = {Graph-based semi-supervised learning: A comprehensive review},
  year      = {2022},
  number    = {11},
  pages     = {8174--8194},
  volume    = {34},
  publisher = {IEEE},
}

@Article{anusree2025understanding,
  author    = {Anusree, M and Pramod, P Nair},
  journal   = {Nonlinear Dynamics},
  title     = {Understanding chaotic neural networks: A comprehensive review},
  year      = {2025},
  pages     = {1--16},
  publisher = {Springer},
}

@Article{balakrishnan2019chaosnet,
  author    = {Balakrishnan, Harikrishnan Nellippallil and Kathpalia, Aditi and Saha, Snehanshu and Nagaraj, Nithin},
  journal   = {Chaos: An Interdisciplinary Journal of Nonlinear Science},
  title     = {ChaosNet: A chaos based artificial neural network architecture for classification},
  year      = {2019},
  number    = {11},
  volume    = {29},
  publisher = {AIP Publishing},
}

@Article{sethi2023neurochaos,
  author    = {Sethi, Deeksha and Nagaraj, Nithin and others},
  journal   = {Integration},
  title     = {Neurochaos feature transformation for Machine Learning},
  year      = {2023},
  pages     = {157--162},
  volume    = {90},
  publisher = {Elsevier},
}

@Article{harikrishnan2021noise,
  author    = {Harikrishnan, Nellippallil Balakrishnan and Nagaraj, Nithin},
  journal   = {Neural Networks},
  title     = {When noise meets chaos: Stochastic resonance in neurochaos learning},
  year      = {2021},
  pages     = {425--435},
  volume    = {143},
  publisher = {Elsevier},
}

@Article{zhou2005tri,
  author    = {Zhou, Zhi-Hua and Li, Ming},
  journal   = {IEEE Transactions on knowledge and Data Engineering},
  title     = {Tri-training: Exploiting unlabeled data using three classifiers},
  year      = {2005},
  number    = {11},
  pages     = {1529--1541},
  volume    = {17},
  publisher = {IEEE},
}

@Article{bennett1998semi,
  author  = {Bennett, Kristin and Demiriz, Ayhan},
  journal = {Advances in Neural Information processing systems},
  title   = {Semi-supervised support vector machines},
  year    = {1998},
  volume  = {11},
}

@Article{mallapragada2008semiboost,
  author    = {Mallapragada, Pavan Kumar and Jin, Rong and Jain, Anil K and Liu, Yi},
  journal   = {IEEE transactions on pattern analysis and machine intelligence},
  title     = {Semiboost: Boosting for semi-supervised learning},
  year      = {2008},
  number    = {11},
  pages     = {2000--2014},
  volume    = {31},
  publisher = {IEEE},
}

@Article{tamposis2019semi,
  author    = {Tamposis, Ioannis A and Tsirigos, Konstantinos D and Theodoropoulou, Margarita C and Kontou, Panagiota I and Bagos, Pantelis G},
  journal   = {Bioinformatics},
  title     = {Semi-supervised learning of Hidden Markov Models for biological sequence analysis},
  year      = {2019},
  number    = {13},
  pages     = {2208--2215},
  volume    = {35},
  publisher = {Oxford University Press},
}

@InProceedings{du2021self,
  author    = {Du, Jingfei and Grave, Edouard and Gunel, Beliz and Chaudhary, Vishrav and Celebi, Onur and Auli, Michael and Stoyanov, Veselin and Conneau, Alexis},
  booktitle = {Proceedings of the 2021 Conference of the North American chapter of the association for computational linguistics: human language technologies},
  title     = {Self-training improves pre-training for natural language understanding},
  year      = {2021},
  pages     = {5408--5418},
}

@InProceedings{ghiasi2021multi,
  author    = {Ghiasi, Golnaz and Zoph, Barret and Cubuk, Ekin D and Le, Quoc V and Lin, Tsung-Yi},
  booktitle = {Proceedings of the IEEE/CVF International Conference on Computer Vision},
  title     = {Multi-task self-training for learning general representations},
  year      = {2021},
  pages     = {8856--8865},
}

@Article{HENRY2025117213,
  author  = {Akhila Henry and Nithin Nagaraj},
  journal = {Chaos, Solitons and Fractals},
  title   = {Augmented regression models using neurochaos learning},
  year    = {2025},
  issn    = {0960-0779},
  pages   = {117213},
  volume  = {201},
}

@Article{as2023analysis,
  author    = {AS, Remya Ajai and Harikrishnan, Nellippallil Balakrishnan and Nagaraj, Nithin},
  journal   = {Chaos, Solitons and Fractals},
  title     = {Analysis of logistic map based neurons in neurochaos learning architectures for data classification},
  year      = {2023},
  pages     = {113347},
  volume    = {170},
  publisher = {Elsevier},
}

@Article{as7random,
  author    = {AS, Remya Ajai and Nagaraj, Nithin},
  journal   = {Chaos Theory and Applications},
  title     = {Random Heterogeneous Neurochaos Learning Architecture for Data Classification},
  number    = {1},
  pages     = {10--30},
  volume    = {7},
  publisher = {Akif AKG{\"U}L},
}

@Article{henry2025neurochaos,
  author    = {Henry, Akhila and Nagaraj, Nithin},
  journal   = {Chaos Theory and Applications},
  title     = {Neurochaos learning for classification using composition of chaotic maps},
  year      = {2025},
  number    = {2},
  pages     = {107--116},
  volume    = {7},
  publisher = {Akif AKG{\"U}L},
}

@Article{pant2025advancing,
  author  = {Pant, Kunal Kumar and Nagaraj, Nithin and others},
  journal = {arXiv preprint arXiv:2510.26383},
  title   = {Advancing Forest Fires Classification using Neurochaos Learning},
  year    = {2025},
}

@InProceedings{anusree2024hypothetical,
  author       = {Anusree, M and Reshmi, P and Valadi, Jayaram and Nair, Pramod P and Suravajhala, Prashanth},
  booktitle    = {International Conference on Information and Communication Technology for Competitive Strategies},
  title        = {Hypothetical Protein Classification Using NeuroChaos Learning Architecture},
  year         = {2024},
  organization = {Springer},
  pages        = {337--346},
}

@Article{harikrishnan2022classification,
  author    = {Harikrishnan, NB and Pranay, SY and Nagaraj, Nithin},
  journal   = {Medical \& Biological Engineering \& Computing},
  title     = {Classification of SARS-CoV-2 viral genome sequences using Neurochaos Learning},
  year      = {2022},
  number    = {8},
  pages     = {2245--2255},
  volume    = {60},
  publisher = {Springer},
}

@Article{nb2022causality,
  author  = {NB, Harikrishnan and Kathpalia, Aditi and Nagaraj, Nithin},
  journal = {Advances in Neural Information Processing Systems},
  title   = {Causality preserving chaotic transformation and classification using neurochaos learning},
  year    = {2022},
  pages   = {2046--2058},
  volume  = {35},
}

@Article{henry2025simplified,
  author    = {Henry, Akhila and Sundaravaradhan, Rajan and Nagaraj, Nithin},
  journal   = {Chaos: An Interdisciplinary Journal of Nonlinear Science},
  title     = {Simplified neurochaos learning architectures for data classification},
  year      = {2025},
  number    = {6},
  volume    = {35},
  publisher = {AIP Publishing},
}

@InProceedings{sneha2023biologically,
  author       = {Sneha, KH and Sudeesh, Adhithya and Nair, Pramod P and Suravajhala, Prashanth},
  booktitle    = {2023 Fifth International Conference on Electrical, Computer and Communication Technologies (ICECCT)},
  title        = {Biologically inspired chaosnet architecture for hypothetical protein classification},
  year         = {2023},
  organization = {IEEE},
  pages        = {1--6},
}

@InProceedings{kahn2020self,
  author       = {Kahn, Jacob and Lee, Ann and Hannun, Awni},
  booktitle    = {ICASSP 2020-2020 IEEE International Conference on Acoustics, Speech and Signal Processing (ICASSP)},
  title        = {Self-training for end-to-end speech recognition},
  year         = {2020},
  organization = {IEEE},
  pages        = {7084--7088},
}

@Article{scudder1965adaptive,
  author    = {Scudder, H},
  journal   = {IEEE Transactions on Information Theory},
  title     = {Adaptive communication receivers},
  year      = {1965},
  number    = {2},
  pages     = {167--174},
  volume    = {11},
  publisher = {IEEE},
}

@Article{wang2022freematch,
  author  = {Wang, Yidong and Chen, Hao and Heng, Qiang and Hou, Wenxin and Fan, Yue and Wu, Zhen and Wang, Jindong and Savvides, Marios and Shinozaki, Takahiro and Raj, Bhiksha and others},
  journal = {arXiv preprint arXiv:2205.07246},
  title   = {Freematch: Self-adaptive thresholding for semi-supervised learning},
  year    = {2022},
}

@Article{feofanov2024multi,
  author  = {Feofanov, Vasilii and Devijver, Emilie and Amini, Massih-Reza},
  journal = {Journal of Machine Learning Research},
  title   = {Multi-class probabilistic bounds for majority vote classifiers with partially labeled data},
  year    = {2024},
  number  = {104},
  pages   = {1--47},
  volume  = {25},
}

@Article{zhang2021flexmatch,
  author  = {Zhang, Bowen and Wang, Yidong and Hou, Wenxin and Wu, Hao and Wang, Jindong and Okumura, Manabu and Shinozaki, Takahiro},
  journal = {Advances in neural information processing systems},
  title   = {Flexmatch: Boosting semi-supervised learning with curriculum pseudo labeling},
  year    = {2021},
  pages   = {18408--18419},
  volume  = {34},
}

@Article{tur2005combining,
  author    = {Tur, Gokhan and Hakkani-T{\"u}r, Dilek and Schapire, Robert E},
  journal   = {Speech Communication},
  title     = {Combining active and semi-supervised learning for spoken language understanding},
  year      = {2005},
  number    = {2},
  pages     = {171--186},
  volume    = {45},
  publisher = {Elsevier},
}

@InProceedings{zou2018unsupervised,
  author    = {Zou, Yang and Yu, Zhiding and Kumar, BVK and Wang, Jinsong},
  booktitle = {Proceedings of the European conference on computer vision (ECCV)},
  title     = {Unsupervised domain adaptation for semantic segmentation via class-balanced self-training},
  year      = {2018},
  pages     = {289--305},
}

@InProceedings{harikrishnan2020neurochaos,
  author       = {Harikrishnan, NB and Nagaraj, Nithin},
  booktitle    = {2020 International Conference on Signal Processing and Communications (SPCOM)},
  title        = {Neurochaos inspired hybrid machine learning architecture for classification},
  year         = {2020},
  organization = {IEEE},
  pages        = {1--5},
}

\end{document}